\documentclass[10pt,twocolumn,letterpaper]{article}

\usepackage[pagenumbers]{cvpr} 

\definecolor{mylightgray}{gray}{0.92}

\newcommand{\cmark}{\ding{51}}%
\newcommand{\xmark}{\ding{55}}%

\newcommand{\method}{Show or Tell}
\newcommand{\short}{SoT}
\newcommand{\myparagraph}[1]{\vspace{0.1cm}\noindent\textbf{#1}\ }

\definecolor{cvprblue}{rgb}{0.21,0.49,0.74}
\usepackage[pagebackref,breaklinks,colorlinks,allcolors=cvprblue]{hyperref}
\usepackage{graphicx}
\usepackage{booktabs}
\usepackage{multirow}
\usepackage{amssymb}
\usepackage{pifont}
\usepackage{algorithm}
\usepackage{algpseudocode}
\usepackage{dsfont}
\usepackage{amsmath}
\usepackage{mwe}
\usepackage{color, colortbl}
\usepackage[accsupp]{axessibility} 

\expandafter\def\expandafter\normalsize\expandafter{%
    \normalsize%
    \setlength\abovedisplayskip{2pt}%
    \setlength\belowdisplayskip{2pt}%
    \setlength\abovedisplayshortskip{2pt}%
    \setlength\belowdisplayshortskip{2pt}%
}

\def\paperID{21} 
\def\confName{CVPR}
\def\confYear{2025}

\title{Show or Tell? A Benchmark To Evaluate Visual and Textual Prompts in Semantic Segmentation}

\author{Gabriele Rosi$^{1,2}$, Fabio Cermelli$^2$ \\
$^1$ Politecnico di Torino, $^2$ Focoos AI\\ 
$^1$ {\tt\small name.surname@polito.it}, $^2$ {\tt\small name.surname@focoos.ai}
}

\begin{document}
\maketitle
\begin{abstract}
Prompt engineering has shown remarkable success with large language models, yet its systematic exploration in computer vision remains limited. In semantic segmentation, both textual and visual prompts offer distinct advantages: textual prompts through open-vocabulary methods allow segmentation of arbitrary categories, while visual reference prompts provide intuitive reference examples. However, existing benchmarks evaluate these modalities in isolation, without direct comparison under identical conditions. We present \method\ (\short), a novel benchmark specifically designed to evaluate both visual and textual prompts for semantic segmentation across 14 datasets spanning 7 diverse domains (common scenes, urban, food, waste, parts, tools, and land-cover). We evaluate 5 open-vocabulary methods and 4 visual reference prompt approaches, adapting the latter to handle multi-class segmentation through a confidence-based mask merging strategy. 
Our extensive experiments reveal that open-vocabulary methods excel with common concepts easily described by text but struggle with complex domains like tools while visual reference prompt methods achieve good average results but exhibit high variability depending on the input prompt. 
Through comprehensive quantitative and qualitative analysis, we identify the strengths and weaknesses of both prompting modalities, providing valuable insights to guide future research in vision foundation models for segmentation tasks. Code is available at \href{https://github.com/FocoosAI/ShowOrTell}{https://github.com/FocoosAI/ShowOrTell}.
\end{abstract}
\vspace{-1.3em}
\section{Introduction}
\label{sec:intro}
A long-standing goal of artificial intelligence is to create models that can generalize and adapt to multiple tasks without requiring complex and costly fine-tuning or retraining~\cite{thrun1998learning, finn2017model, kirkpatrick2017overcoming, zenke2017continual, parisi2019continual, li2017learning, cermelli2020modeling}.
Prompt engineering has recently revolutionized large language models (LLMs) showing that optimizing input instructions can substantially alter model behavior and improve performance~\cite{chen2023unleashing}. However, systematic investigations into prompt engineering for computer vision tasks remain in their early stages~\cite{wang2023review, kirillov2023segment}.

\begin{figure}[t]
  \centering
  \includegraphics[width=0.85\linewidth]{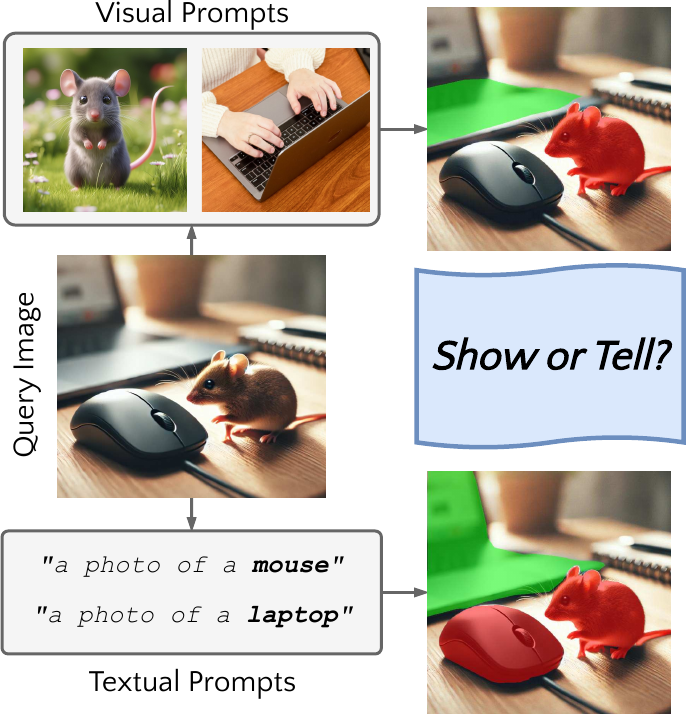}
  \caption{Our \method\ (\short) benchmark evaluates the effectiveness of textual and visual prompts in semantic segmentation. Here the query image poses a challenge for textual prompts since ``mouse'' can refer to both an animal and a computer accessory, resulting in an inaccurate segmentation. Similarly, visual prompts also face difficulties in accurately identifying the ``laptop" due to the limited information provided by a top-down view.}
  \label{fig:teaser}
\end{figure}

Recent work in computer vision has begun exploring how insights from prompt engineering can enhance semantic segmentation. By integrating carefully designed prompts into segmentation pipelines, researchers are uncovering new ways to guide pixel-level predictions and improve model flexibility.
Pioneering research in this direction are represented by open-vocabulary semantic segmentation~\cite{ghiasi2022scaling, barsellotti2023enhancing, liang2023open, qin2023freeseg, cha2023learning, mukhoti2023open} which has initiated the exploration of textual prompting for vision tasks.
These approaches leverage Vision Foundation Models (VFMs), particularly CLIP~\cite{radford2021learning, ilharco2021openclip}, to extract and align textual embeddings from class names with visual features, thereby enabling the segmentation of arbitrary categories without prior training on specific classes. Despite their efficacy, these textual approaches encounter substantial limitations.
Indeed, difficult concepts, such as the various types of cracks on a wall or a specific bird species, are hard to describe with text alone, as textual descriptions may fail to capture the unique visual cues needed for accurate recognition.
Moreover, generic textual prompts frequently fail to distinguish between semantically distinct objects: as shown in~\cref{fig:teaser}, the prompt ``a photo of a mouse" results in the segmentation of both the animal and the computer peripheral, highlighting the need for more precise contextual information.

To address these limitations, visual prompts have emerged as a compelling alternative due to their intuitive nature and straightforward implementation~\cite{kirillov2023segment, ravi2024sam, shtedritski2023does}. Human perception typically processes visual information before linguistic categorization, making visual prompting a natural approach to object identification. The use of masks, bounding boxes, or scribbles to indicate objects of interest provides an effective mean for identifying concepts that resist precise verbal description. Visual reference prompt methods~\cite{kirillov2023segment,liu2023matcher,zhang2024bridge,sun2024vrp,zhang2023personalize} have yielded significant advancements in image segmentation. Recent approaches~\cite{liu2023matcher, zhang2024bridge} typically employ a two-stage pipeline leveraging VFMs: initially, a model such as DINO~\cite{caron2021emerging, oquab2023dinov2} identifies pixels exhibiting similarity to the reference class, followed by prompting SAM~\cite{kirillov2023segment} to generate the segmentation masks. However, relying on visual prompts poses challenges when objects vary in appearance across instances.
As depicted in \cref{fig:teaser} choosing the right visual prompts could be challenging since they might contain limited visual cues, thus reducing the segmentation performance.

Based on these observations, each modality, textual and visual, offers distinct advantages and limitations. Comparing them provides two important benefits: it reveals scenarios where one type of prompt performs better than the other, and it helps determine which modality works best for semantic segmentation tasks. Notably, existing benchmarks~\cite{zou2023generalized, blumenstiel2023mess} typically evaluate visual and textual prompts separately, without directly comparing them under the same conditions. Additionally, previous studies have tested visual reference prompt methods only for single-class segmentation~\cite{sun2024vrp, liu2023matcher, zhang2024bridge}, failing to assess their performance in complex real-world scenarios.

To bridge this gap, we introduce a novel benchmark, \textit{\method\ (\short)}, specifically designed to evaluate both visual and textual prompts within the context of semantic segmentation. To properly assess the generalization ability, \short\ spans 14 datasets across 7 domains (common, urban, food, waste, parts, tools, and land-cover). This benchmark enables a direct, head-to-head evaluation of different prompting methods in semantic segmentation, assessing their adaptability by only changing the prompts. To the best of our knowledge, this is the first effort toward an evaluation of multiple prompting techniques in semantic segmentation.

Our evaluation encompasses five open-vocabulary methods (MaskCLIP~\cite{chen2023exploring}, TCL~\cite{liu2022learning}, CLIP-DINoiser~\cite{wysoczanska2024clip}, NACLIP~\cite{hajimiri2024pay}, and ProxyCLIP~\cite{lan2024proxyclip}) and four visual reference prompt approaches (SINE~\cite{liu2024simple}, PerSAM~\cite{zhang2023personalize}, Matcher~\cite{liu2023matcher}, and GFSAM~\cite{zhang2024bridge}), all assessed under identical experimental conditions. To facilitate a fair comparison of visual reference prompt methods, which were originally designed for binary segmentation tasks, we implement a novel adaptation for multi-class segmentation scenarios. This adaptation involves generating individual binary masks for each class and subsequently merging them according to the model's confidence scores, enabling these methods to effectively segment scenes containing multiple objects.

Extensive experimental results highlight that open-vocabulary methods excel in domains where the classes represent common concepts that can be easily described by text, while visual reference prompt methods obtain good results on average but their results can significantly differ depending on the input prompt.

In summary, the contributions of the paper are three-fold:
\begin{itemize}
  \item We introduce a novel benchmark, \short, comprising 14 datasets in 7 domains and systematically comparing 4 visual and 5 textual prompting methods;
  \item We adapt visual reference prompt methods to operate in multi-class setting;
  \item Through an extensive quantitative and qualitative analysis, we identify the strengths and weaknesses of both prompting modalities, providing valuable insights for future research.
\end{itemize}

\section{Related works}
\label{sec:related}

\myparagraph{Prompt engineering for semantic segmentation.}
Prompt engineering, originating in natural language processing, has expanded to computer vision applications. Initially focused on text-to-image generation with models like DALL-E~\cite{ramesh2021zero} and Stable Diffusion~\cite{rombach2022high}, this approach evolved with vision-language models such as CLIP~\cite{radford2021learning}, which demonstrated how textual prompts could shape visual representations for zero-shot recognition.
In semantic segmentation, prompting techniques are used to guide foundation models including CLIP~\cite{radford2021learning}, DINO~\cite{caron2021emerging, oquab2023dinov2}, and SAM~\cite{kirillov2023segment}. Most approaches rely exclusively on either text prompts (Open-Vocabulary Semantic Segmentation~\cite{guo2023mvp, lan2024proxyclip, mukhoti2023open, wysoczanska2024clip}) or visual prompts (Visual Reference Prompt Segmentation~\cite{sun2024vrp,liu2023matcher,zhang2024bridge}), with few exceptions~\cite{luddecke2022image, nagendra2024samic} combining both modalities.
While previous benchmarks~\cite{zou2023generalized, blumenstiel2023mess} typically evaluate the approaches separately, our work directly compares them to assess their respective strengths and weaknesses in real-world scenarios.

\myparagraph{Open-vocabulary segmentation.}
Open-vocabulary semantic segmentation extends traditional zero-shot semantic segmentation~\cite{xian2019semantic, bucher2019zero, gu2020context, pastore2021closer} by enabling models to identify arbitrary object categories without class-specific training, leveraging the semantic understanding capabilities of vision-language foundation models~\cite{radford2021learning, caron2021emerging, oquab2023dinov2, kirillov2023segment} to generalize beyond predefined class sets.
The extension of CLIP's aligned image-language representations~\cite{radford2021learning,ilharco2021openclip} to segmentation presents challenges, as CLIP's architecture is not inherently designed for dense vision-language features.
Initial approaches relied on fine-tuning CLIP to semantic segmentation by exploiting a large labeled set containing pixel-wise annotations from a large set of classes~\cite{ghiasi2022scaling, barsellotti2023enhancing, liang2023open, qin2023freeseg, cha2023learning, mukhoti2023open}. However, these approaches require a large amount of annotated data and they alter the open-vocabulary capabilities of the original CLIP model, biasing its knowledge on the training dataset.
For this reason, recent works have focused on training-free approaches~\cite{zhou2022extract, wang2024sam, wysoczanska2024clip, lan2024proxyclip, hajimiri2024pay, shin2022reco, karazija2023diffusion}.
Recent innovations include MaskCLIP~\cite{zhou2022extract}, which revealed that value embeddings offer better localization than token embeddings, and approaches that refine CLIP's attention mechanisms~\cite{li2023clip, wang2024sclip, hajimiri2024pay}. Other methods~\cite{wysoczanska2024clip, wang2024sam, lan2024proxyclip} combine CLIP's semantic understanding with spatial consistency from models like DINO and SAM.

\myparagraph{Visual Reference Segmentation.}
Visual Reference Segmentation employs annotated reference images to guide the segmentation of semantically similar regions in target images. Originating as Few-Shot Segmentation (FSS)~\cite{shaban2017one}, early approaches~\cite{wang2019panet, zhang2019pyramid, li2021adaptive, cermelli2020prototype} concentrated on training neural networks to extract prototypes from reference images and compute similarity for target image segmentation. Due to the limitations of prototype-based methods, subsequent research proposed extracting correlation maps to represent the relationship between reference and target images~\cite{min2021hypercorrelation, wang2020few} or utilizing attention maps to guide target image segmentation~\cite{zhang2021few, hong2022cost}.
The advent of vision foundation models (VFMs) has transformed the field, directing research toward the use of large-scale pre-trained models for target image segmentation. Several approaches~\cite{wang2023images, wang2023seggpt, liu2024simple, xiao2024cat} have developed models with cross-task generalization capabilities. Painter~\cite{wang2023images} introduced an in-context learning framework wherein vision tasks are defined through exemplars. SINE~\cite{liu2024simple} presented an encoder-decoder architecture that handles multiple tasks via in-context examples. However, the training of such models requires substantial computational resources and extensive datasets. Consequently, research has increasingly focused on employing existing VFMs such as DINO~\cite{caron2021emerging, oquab2023dinov2} and SAM~\cite{kirillov2023segment, ravi2024sam} in a training-free manner, as these models offer superior generalization capabilities due to their comprehensive pretraining. PerSAM~\cite{zhang2023personalize} adapts SAM for personalized segmentation with minimal parameter modifications. VRP-SAM~\cite{sun2024vrp} incorporates SAM with an external feature-matching encoder without fine-tuning. Matcher~\cite{liu2023matcher} and GFSAM~\cite{zhang2024bridge} implement two-stage pipelines that employ DINOv2~\cite{oquab2023dinov2} to compute cross-image similarities for extracting prompts for SAM.

\section{\method\ Benchmark}
\label{sec:method}
To facilitate a rigorous comparison between visual and textual prompts for semantic segmentation, we introduce \textit{\method\ (\short)}. This novel benchmark evaluates 5 open-vocabulary and 4 visual reference prompt methods across 14 distinct datasets spanning 7 domains.
In the following sections, we present an overview of the methods included in the benchmark (\cref{subsec:models}) and describe our approach for adapting visual reference prompts to generate predictions for multi-class semantic segmentation (\cref{subsec:adapting}). Subsequently, we describe the composition of the benchmark, including the datasets and their challenges (\cref{subsec:datasets}).


\begin{table}
    \centering
    \setlength{\tabcolsep}{2pt}
    \renewcommand{\arraystretch}{1.2}
    \resizebox{\columnwidth}{!}{
    \begin{tabular}{lccccc} 
        \toprule
         Name& Prompt type& Visual backbone& Trained \\ 
         \hline
         SINE \cite{liu2024simple} \scriptsize{[\textit{NeurIPS'24}]}               & \multirow{4}{*}{Visual } &  DINOv2 \cite{oquab2023dinov2}         & \cmark  \\
         PerSAM \cite{zhang2023personalize} \scriptsize{[\textit{ICLR'24}]}            &                         &  SAM \cite{kirillov2023segment}           & \xmark \\
         Matcher \cite{liu2023matcher} \scriptsize{[\textit{ICLR'24}]}          &                         &  DINOv2 \cite{oquab2023dinov2}  + SAM  \cite{kirillov2023segment}  & \xmark  \\
         GFSAM \cite{zhang2024bridge} \scriptsize{[\textit{NeurIPS'24}]}  &                         &  DINOv2 \cite{oquab2023dinov2}  + SAM  \cite{kirillov2023segment}  & \xmark  \\ 
        \midrule
         TCL \cite{cha2023learning} \scriptsize{[\textit{CVPR'23}]}                 & \multirow{5}{*}{Textual}&  CLIP \cite{radford2021learning}   & \cmark   \\ 
         MaskCLIP \cite{chen2023exploring} \scriptsize{[\textit{ECCV'22}]}           &                         &  CLIP \cite{radford2021learning}    & \xmark   \\ 
         CLIP-DINoiser \cite{wysoczanska2024clip} \scriptsize{[\textit{ECCV'24}]}       &                         &  CLIP \cite{radford2021learning}    & \xmark   \\ 
         NACLIP \cite{hajimiri2024pay} \scriptsize{[\textit{WACV'25}]}             &                         &  CLIP \cite{radford2021learning}    & \xmark   \\ 
         ProxyCLIP \cite{lan2024proxyclip} \scriptsize{[\textit{ECCV'24}]}           &                         &  CLIP \cite{radford2021learning}  + DINO \cite{caron2021emerging, oquab2023dinov2}  & \xmark   \\ 
        \bottomrule
    \end{tabular}}
    \caption{List of models analyzed in our benchmark. For each model we report the type of prompt employed, the visual backbone(s) used and if the model undergone a training process.}
    \label{tab:models}
\end{table}

\subsection{Models}
\label{subsec:models}
A comprehensive evaluation of visual and textual prompts requires the selection of an appropriate subset of models for each prompt category. To accomplish this objective, we first conduct an analysis of the operational mechanisms of models employing these prompt types, followed by a justification of our selection criteria. A summary of the benchmarked methods is presented in \cref{tab:models}.

\myparagraph{Textual prompting based.} Text serves as a powerful prompt, enabling the description of visual concepts using natural language. In recent years, this approach has been widely adopted by open-vocabulary segmentation (OVSS) models. Specifically, in the context of OVSS, the goal is to assign a semantic label (provided as a free-form text description) to each pixel (or region) in an image, without relying on a predefined set of labels.
OVSS methods operate on an image and a vocabulary of textual class descriptions. These approaches extract dense visual features from the image using a visual encoder while simultaneously generating embeddings for each class in the vocabulary via a text encoder. The core challenge lies in aligning these visual and textual representations to produce accurate semantic segmentation maps for all prompted classes.

For our benchmark, we evaluate five representative methods: MaskCLIP~\cite{zhou2022extract}, TCL~\cite{cha2023learning}, CLIP-DINoiser~\cite{wysoczanska2024clip}, NACLIP~\cite{hajimiri2024pay}, and ProxyCLIP~\cite{lan2024proxyclip}. We deliberately emphasize training-free approaches that leverage VFMs like DINO~\cite{caron2021emerging, oquab2023dinov2} and CLIP~\cite{radford2021learning} rather than models specifically trained for segmentation tasks. This selection strategy serves dual purposes: it prevents potential bias from dataset contamination that might favor certain domains, and it better aligns with our cross-domain evaluation objectives. Training-free methods typically demonstrate stronger generalization capabilities due to their diverse pretraining on large-scale datasets~\cite{radford2021learning, ilharco2021openclip}, allowing us to assess their intrinsic ability to transfer knowledge to novel visual concepts without task-specific optimization.

\myparagraph{Visual prompting based.} Visual prompts represent concepts through explicit visual cues such as masks, bounding boxes, or scribbles, offering an alternative to textual descriptions. These prompts are predominantly employed in few-shot segmentation (FSS)~\cite{shaban2017one, liu2023matcher, sun2024vrp}, where the objective is to generate a segmentation mask for a query image that identifies the same class annotated in a small support set of images.
While FSS methods demonstrate adaptability to novel classes, they typically operate in a single-class segmentation paradigm. Although some approaches have been extended to generalized and incremental few-shot segmentation~\cite{xian2019semantic, cermelli2020prototype}, these methods are trained on specific datasets (\eg, PASCAL VOC~\cite{everingham2010pascal}, ADE20K~\cite{zhou2017scene}) and lack the generalization capabilities required for our evaluation.

For our benchmark, we specifically selected visual reference prompt methods that leverage vision foundation models (VFMs) such as DINO~\cite{caron2021emerging, oquab2023dinov2} and SAM~\cite{kirillov2023segment}, prioritizing their inherent generalizability across diverse visual domains. The benchmark includes SINE~\cite{liu2024simple}, PerSAM~\cite{zhang2023personalize}, Matcher~\cite{liu2023matcher}, and GFSAM~\cite{zhang2024bridge}. Among these, SINE~\cite{liu2024simple} employs DINO~\cite{oquab2023dinov2} for visual prompting, and fine-tunes a transformer decoder to produce mask at different granularity able to perform in-context segmentation.
Despite their effectiveness, these methods are designed to segment only one class at a time, necessitating adaptation for multi-class semantic segmentation in our benchmark.

\begin{table}[t]
    \centering
    \setlength{\tabcolsep}{2pt}
    \renewcommand{\arraystretch}{1.2}
    \resizebox{\columnwidth}{!}{
    \begin{tabular}{ccccc} 
        \toprule
        Domain & Dataset & Classes & Train images & Val. images \\ 
        \hline
        \multirow{2}{*}{\textbf{Common}} 
        & ADE20K \cite{zhou2017scene} & 150  &  20,000&  2,000\\
        & PASCAL VOC 2012 \cite{everingham2010pascal} & 21 & 1,464 & 1,449 \\         \hline

        \multirow{2}{*}{\textbf{Urban}}
         & Cityscapes \cite{cordts2016cityscapes} & 19  & 2,975 &  500\\ 
         & UAVid \cite{lyu2020uavid}                  & 7  &  200    &  70\\         \hline

        \multirow{2}{*}{\textbf{Waste}}
         & Trash \cite{trash-dataset}                   & 12   &  832     &  92\\
         & ZeroWaste \cite{bashkirova2022zerowaste}     & 4   &  3,002   &  572 \\         \hline

         \multirow{2}{*}{\textbf{Food}}
         & Pizza \cite{pizza_dataset}                    & 5    &  437     &  122\\
         & UECFood \cite{ege2019new}    & 102   &  9,000   &  1,000\\         \hline

         \multirow{2}{*}{\textbf{Tools}}
         & Toolkits \cite{toolkit-dataset}     & 8   &  48      &  6\\
         & PIDray \cite{zhang2023pidray}                & 12  &  29,454  &  3,733\\         \hline

         \multirow{2}{*}{\textbf{Parts}}
         & House-Parts \cite{house-dataset}              & 22  &  700     &  201\\
         & MHPv1 \cite{li2017multiple}                  & 17 &  4,000   &  980\\         \hline

         \multirow{2}{*}{\textbf{Land-Cover}}
         & LoveDA-Rural \cite{wang2021loveda}                & 6   &  1,366     &  992\\
         & LoveDA-Urban \cite{wang2021loveda}                & 6   &  1,156     &  667\\
        \bottomrule
    \end{tabular}}
    \caption{Our benchmark \short\ is composed of 14 datasets divided into 7 different domains. For each dataset, we report the number of classes and the number of train and validation images.}
    \label{tab:datasets}
\end{table}

\subsection{Adapting Visual Reference Prompt Methods}
\label{subsec:adapting}
Visual reference prompt methods are designed to generate a single mask when provided with an annotated support set. These methods accept a support set containing images with masked annotations of a target class and subsequently produce a binary segmentation map identifying the corresponding class in a query image. For comprehensive evaluation in \short, however, these methods require adaptation to accommodate multi-class scenarios. We implement this adaptation through a two-stage process: first, computing individual masks for each class, and second, integrating these masks based on the model's confidence for each prediction.

\myparagraph{Producing multiple masks.}
To adapt visual reference prompt methods for multi-class segmentation, we first create a semantic support set $\mathcal{S}^{sem}$ by randomly sampling $k$ images for each of the $n$ classes in the dataset . This results in a total of $n \times k$ support images. When processing a query image $q$, we run the model $n$ separate times, once for each class, to generate a binary mask for each class. While this approach is straightforward to implement, it requires $n$ separate forward passes, making it computationally expensive. We analyze this efficiency issue in detail in \cref{subsec:computation}.

\myparagraph{Merging predictions.} After processing a query image, visual reference prompt methods produce $n$ independent binary masks, one for each class. To create a unified semantic segmentation map, we must assign a confidence score to each mask. We standardize this process across methods: for training-free approaches (such as PerSAM~\cite{zhang2023personalize}, Matcher~\cite{liu2023matcher}, and GFSAM~\cite{zhang2024bridge}), we use the mean confidence score from the visual backbone (DINOv2~\cite{oquab2023dinov2} or SAM~\cite{kirillov2023segment}) within each predicted mask. For trained methods like SINE~\cite{liu2024simple}, we use the output probability of the decoder. Once these confidence scores are assigned, we apply an argmax operation across all class masks to produce the final semantic segmentation prediction.


\begin{figure}[ht]
  \centering
  \includegraphics[width=\columnwidth]{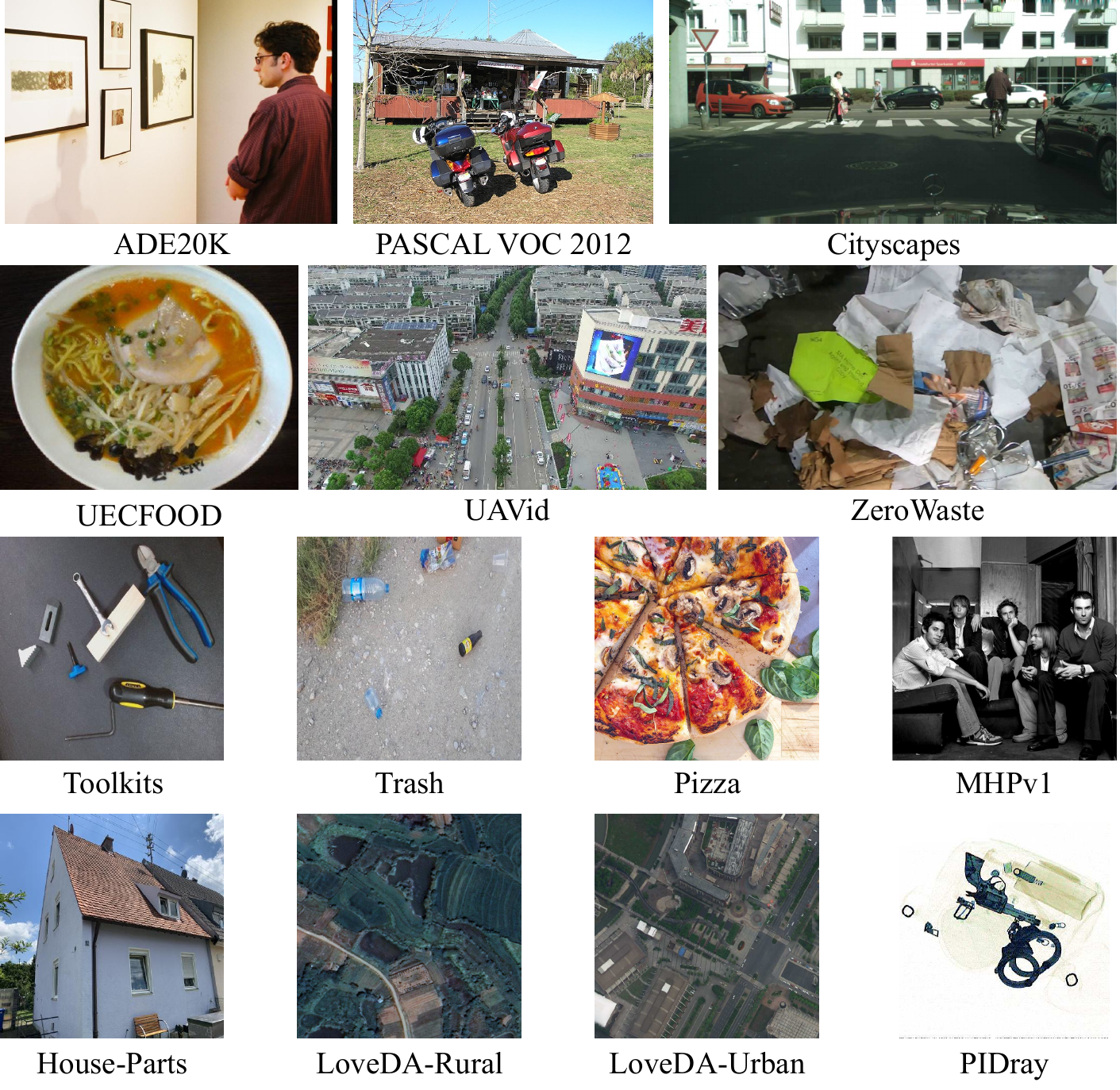}
  \caption{Sample images drawn from the datasets that compose our \short\ benchmark.}
  \label{fig:dataset_examples}
\end{figure}

\subsection{Datasets}
\label{subsec:datasets}
Our benchmark, \short, aims to rigorously evaluate how textual and visual prompting methods perform across diverse real-world scenarios. To achieve this, we carefully selected 14 datasets spanning 7 distinct domains. This selection was motivated by the need to assess these methods beyond controlled environments, following recent benchmarks that test segmentation approaches ``in the wild"~\cite{zou2023generalized, blumenstiel2023mess}. We deliberately included both common scenes and specialized applications to ensure comprehensive coverage of visual contexts that practitioners might encounter. In the following, we describe each domain and its constituent datasets, emphasizing their unique visual characteristics that might differently affect the performance of textual versus visual prompting approaches. \Cref{tab:datasets} provides a summary of all datasets included in our benchmark, while \Cref{fig:dataset_examples} reports some example images from each dataset.

\myparagraph{Common} scenes represent foundational benchmarks in segmentation research. ADE20K~\cite{zhou2017scene} encompasses 150 classes categorized as stuff (\eg \textit{sky, grass, road}) or things (\eg \textit{car, person, chair}), providing a comprehensive framework. In contrast, PASCAL VOC 2012~\cite{everingham2010pascal} adopts an object-focused approach with 20 thing classes distributed across varied indoor and outdoor environments.

\myparagraph{Urban} scenes are critical for autonomous vehicles. Cityscapes~\cite{cordts2016cityscapes} provides high-resolution street-level imagery with 19 classes with dynamic objects and varying weather conditions. UAVid~\cite{lyu2020uavid} captures aerial perspectives with 7 categories, introducing complexities due to scale variations, occlusions, and perspective distortions.

\myparagraph{Waste} domain datasets capture litter in diverse environments. Trash~\cite{trash-dataset} and ZeroWaste~\cite{bashkirova2022zerowaste} contain 12 and 4 waste classes respectively. The variability in shape makes certain classes particularly challenging: for instance, \textit{Styrofoam pieces} can resemble plastic and appear in unpredictable forms, complicating both textual and visual prompting.

\myparagraph{Food} domain presents challenges due to preparation and presentation variability. Pizza~\cite{pizza_dataset} includes categories for pizza base and 4 dressings, while UECFood~\cite{ege2019new} comprises 102 dishes. Classes like \textit{tomato} may appear whole or crushed, while dishes with similar appearance may be difficult to identify (\eg, \textit{miso soup} vs \textit{chinese soup}).

\myparagraph{Tools} domain is important for security and robotics applications. Toolkit~\cite{toolkit-dataset} contains 8 common tools on a workbench with many occlusions, while PIDray~\cite{zhang2023pidray} presents 12 classes in X-ray scans. Objects like \textit{scissors} or \textit{lighters} are particularly challenging to identify in X-ray imagery due to material composition variations and occlusions.

\myparagraph{Parts} segmentation demands understanding of object structures. House-Parts~\cite{house-dataset} includes 22 architectural elements where features like \textit{wooden doors} may lack visual distinction. MHPv1~\cite{li2017multiple} contains 17 human body parts, introducing complexity in distinguishing spatially related elements such as \textit{right leg} versus \textit{left shoe}.

\myparagraph{Land-Cover} applications are increasingly important for satellite imagery analysis. LoveDA~\cite{wang2021loveda} presents a dataset with 6 classes distinguishing rural and urban areas. This domain is challenging as classes like \textit{buildings} are not usually seen from above, while \textit{agricultural} terrain varies significantly based on crop types and seasonal changes.

\section{Results}
\label{sec:results}

\begin{table*}[ht]
    \setlength{\tabcolsep}{3.5pt}
    \renewcommand{\arraystretch}{1.1}
    \centering
    \resizebox{\textwidth}{!}{
    \begin{tabular}{l c c c c c c c c c}
        \toprule
        \multirow{2}{*}{MODEL} & \multirow{2}{*}{Prompt} & \multicolumn{2}{c}{\textit{Common Scene}} & \multicolumn{2}{c}{\textit{Urban}} & \multicolumn{2}{c}{\textit{Waste}} & \multicolumn{2}{c}{\textit{Food}} \\
        \cmidrule(lr){3-4} \cmidrule(lr){5-6} \cmidrule(lr){7-8} \cmidrule(lr){9-10}
        & & PASCAL VOC & ADE20K & Cityscapes & UAVid & Trash & ZeroWaste & Pizza & UECFood \\
       
        \midrule
        \multicolumn{10}{l}{\textbf{Open-vocabulary methods (textual prompt)}} \\
        MaskCLIP \cite{chen2023exploring}  & \textit{Textual} & 38.8 & 9.8 & 12.6 & 23.5 & 4.3 & 6.6 & 25.5 & 4.5 \\
        TCL \cite{cha2023learning}                 & \textit{Textual} & 51.2  & 14.9 & 23.1 & 11.2 & 19.0 & 9.6 & 26.1 & 10.3 \\
        CLIP-DINoiser \cite{wysoczanska2024clip}      & \textit{Textual} & \textbf{62.1} & 20.0 & 31.7 & 25.4 & 13.8 & 14.5 & 37.6 & 13.9 \\
        NACLIP \cite{hajimiri2024pay}              & \textit{Textual} & 52.1 & 17.3 & 31.4 & 25.8 & 18.5 & 10.4 & 43.5 & 16.6 \\
        ProxyCLIP (DINOv2) \cite{lan2024proxyclip} & \textit{Textual} & 58.6 & 21.6 & 35.2 & 29.3 & 18.7 & 14.3 & 49.2 & \underline{22.7} \\
        ProxyCLIP (DINO) \cite{lan2024proxyclip}   & \textit{Textual} & \underline{60.6} & \underline{22.6} & 40.1 & 30.2 & 21.1 & \underline{15.1} & \underline{52.5} & 22.6 \\

         \midrule
        \multicolumn{10}{l}{\textbf{Visual reference prompt methods (visual prompt)}} \\
        
        \multirow{2}{*}{SINE \cite{liu2024simple}} & \textit{1 prompt} & 31.2{\footnotesize $\pm$2.6} & 15.9{\footnotesize $\pm$0.7} & 42.7{\footnotesize $\pm$2.2} & \underline{31.0{\footnotesize $\pm$6.1}} & 9.4{\footnotesize $\pm$1.5} & 8.3{\footnotesize $\pm$0.8} & 15.2{\footnotesize $\pm$1.9} & 1.3{\footnotesize $\pm$0.2} \\
         & \textit{5 prompt} & 37.2{\footnotesize $\pm$1.7} & 18.3{\footnotesize $\pm$0.2} & \underline{44.1{\footnotesize $\pm$1.0}} & \textbf{36.9{\footnotesize $\pm$1.6}} & 14.4{\footnotesize $\pm$2.4} & 8.1{\footnotesize $\pm$1.5} & 19.5{\footnotesize $\pm$2.6} & 2.7{\footnotesize $\pm$0.4} \\
         \multirow{2}{*}{PerSAM \cite{zhang2023personalize}} & \textit{1 prompt} & 10.1{\footnotesize $\pm$1.2} & 2.7{\footnotesize $\pm$0.2} & 15.1{\footnotesize $\pm$0.4} & 6.6{\footnotesize $\pm$1.8} & 3.6{\footnotesize $\pm$1.0} & 3.4{\footnotesize $\pm$1.1} & 16.8{\footnotesize $\pm$1.1} & 1.9{\footnotesize $\pm$0.2} \\
         & \textit{5 prompt} & - & - & - & - & - & - & - & - \\
         \multirow{2}{*}{Matcher \cite{liu2023matcher}} & \textit{1 prompt} & 35.3{\footnotesize $\pm$1.9} & -$^{\dagger}$ & 30.6{\footnotesize $\pm$3.7} & 14.0{\footnotesize $\pm$2.0} & 32.6{\footnotesize $\pm$3.7} & 10.5{\footnotesize $\pm$3.0} & 36.0{\footnotesize $\pm$9.8} & -$^{\dagger}$ \\
         & \textit{5 prompt} & 42.9{\footnotesize $\pm$1.6} & -$^{\dagger}$ & 37.7{\footnotesize $\pm$1.8} & 20.5{\footnotesize $\pm$1.1} & \underline{46.6{\footnotesize $\pm$3.2}} & 13.1{\footnotesize $\pm$1.6}  & 40.2{\footnotesize $\pm$2.8} & -$^{\dagger}$ \\
         \multirow{2}{*}{GFSAM \cite{zhang2024bridge}} & \textit{1 prompt} & 34.0{\footnotesize $\pm$3.6} & 17.4{\footnotesize $\pm$1.0} & 36.5{\footnotesize $\pm$3.3} & 21.7{\footnotesize $\pm$3.8} & 38.6{\footnotesize $\pm$4.4} & 14.2{\footnotesize $\pm$3.4} & 51.6{\footnotesize $\pm$14.1} & 17.9{\footnotesize $\pm$1.4} \\
         & \textit{5 prompt} & 45.0{\footnotesize $\pm$1.4} & \textbf{23.4{\footnotesize $\pm$0.6}} & \textbf{44.6{\footnotesize $\pm$0.9}} & 29.7{\footnotesize $\pm$1.6} & \textbf{52.5{\footnotesize $\pm$5.6}} & \textbf{16.8{\footnotesize $\pm$3.7}} & \textbf{62.2{\footnotesize $\pm$2.7}} & \textbf{26.8{\footnotesize $\pm$1.1}} \\
        
        \bottomrule    
    \end{tabular}}

    \vspace{0.2cm}

    \setlength{\tabcolsep}{3.5pt}
    \renewcommand{\arraystretch}{1.1}
    \centering
    \resizebox{0.95\textwidth}{!}{
    \begin{tabular}{l c c c c c c c >{\columncolor[gray]{0.9}}c}
        \toprule
        \multirow{2}{*}{MODEL} & \multirow{2}{*}{Prompt} & \multicolumn{2}{c}{\textit{Land-Cover}} & \multicolumn{2}{c}{\textit{Tools}} & \multicolumn{2}{c}{\textit{Parts}} & \cellcolor{white} \\
        \cmidrule(lr){3-4} \cmidrule(lr){5-6} \cmidrule(lr){7-8}
        & & LoveDA-Rural & LoveDA-Urban & Toolkits & PIDray & House-Parts & MHPv1 & \textbf{AVG} \\
        
        \midrule
        \multicolumn{9}{l}{\textbf{Open-vocabulary methods (textual prompt)}} \\
        MaskCLIP \cite{chen2023exploring}  & \textit{Textual} & 14.0  & 26.1  & 2.0  & 2.0  & 4.6  & 12.4 & 13.3 \\
        TCL \cite{cha2023learning}                 & \textit{Textual} & 12.3 & 14.9 & 6.5 & 4.8 & 7.8 & 11.5 & 15.9 \\
        CLIP-DINoiser \cite{wysoczanska2024clip}   & \textit{Textual} & 25.3 & 35.3 & 6.0 & 3.5 & 7.3 & 17.3 & 22.4 \\
        NACLIP \cite{hajimiri2024pay}              & \textit{Textual} & 24.2 & 31.1 & 12.9 & 6.4 & 7.4 & 20.9 & 22.8 \\
        ProxyCLIP (DINOv2) \cite{lan2024proxyclip} & \textit{Textual} & 31.4 & \underline{40.3} & 3.4 & 7.4 & 3.4 & 21.5 & 25.5 \\
        ProxyCLIP (DINO) \cite{lan2024proxyclip}   & \textit{Textual} & \underline{32.7} & 35.4 & 6.4 & 7.5 & 3.6 & \underline{28.1} & 27.1 \\

        \midrule
        \multicolumn{9}{l}{\textbf{Visual reference prompt methods (visual prompt)}} \\
        
        \multirow{2}{*}{SINE \cite{liu2024simple}} & \textit{1 prompt} & 14.5{\footnotesize $\pm$5.0} & 26.7{\footnotesize $\pm$4.7} & 43.7{\footnotesize $\pm$8.6} & 1.8{\footnotesize $\pm$0.3} & 12.3{\footnotesize $\pm$1.8} & 8.3{\footnotesize $\pm$1.1}  & 18.7 \\
         & \textit{5 prompt} & 20.3{\footnotesize $\pm$3.9} & 31.9{\footnotesize $\pm$4.2} & 64.0{\footnotesize $\pm$9.2} & 3.2{\footnotesize $\pm$0.3} & 14.7{\footnotesize $\pm$1.7} & 10.0{\footnotesize $\pm$0.5} & 23.2 \\
         \multirow{2}{*}{PerSAM \cite{zhang2023personalize}} & \textit{1 prompt} & 8.1{\footnotesize $\pm$2.1} & 8.7{\footnotesize $\pm$3.0} & 21.7{\footnotesize $\pm$7.5} & 2.3{\footnotesize $\pm$0.4} & 4.0{\footnotesize $\pm$1.6} & 8.3{\footnotesize $\pm$0.8} & 8.1 \\
         & \textit{5 prompt} & - & - & - & - & - & - & - \\
         \multirow{2}{*}{Matcher \cite{liu2023matcher}} & \textit{1 prompt} & 18.3{\footnotesize $\pm$3.1} & 22.7{\footnotesize $\pm$6.2} & 85.6{\footnotesize $\pm$9.7} & 6.4{\footnotesize $\pm$1.8} & 16.0{\footnotesize $\pm$2.4} & 16.3{\footnotesize $\pm$2.0} & 27.2 \\
         & \textit{5 prompt} & 25.4{\footnotesize $\pm$2.6} & 30.5{\footnotesize $\pm$4.8} & \textbf{88.9{\footnotesize $\pm$2.9}} & 10.3{\footnotesize $\pm$1.1} & \textbf{25.0{\footnotesize $\pm$5.0}} & 19.3{\footnotesize $\pm$2.2} & 33.5 \\
         \multirow{2}{*}{GFSAM \cite{zhang2024bridge}} & \textit{1 prompt} & 30.5{\footnotesize $\pm$8.2} & 36.4{\footnotesize $\pm$6.0} & 71.6{\footnotesize $\pm$3.7} & \underline{14.0{\footnotesize $\pm$4.5}} & 18.8{\footnotesize $\pm$3.2} & 21.4{\footnotesize $\pm$1.8} & 30.3 \\
         & \textit{5 prompt} & \textbf{35.6{\footnotesize $\pm$1.4}} & \textbf{43.4{\footnotesize $\pm$3.1}} & \underline{86.9{\footnotesize $\pm$1.4}} & \textbf{21.0{\footnotesize $\pm$2.6}} & \underline{24.7{\footnotesize $\pm$2.6}} & \textbf{28.6{\footnotesize $\pm$1.8}} & 38.7 \\
        
        \bottomrule
    \end{tabular}}
    \caption{Benchmark results for the selected model. 
    We present the mIoU for each model, and for visual reference prompt methods, we also include the standard deviation across different seeds. 
    Note that for PerSAM \cite{zhang2023personalize}, only the configuration with one prompt is reported due to its limitation in handling multiple prompts. $^{\dagger}$ denotes experiments that required excessive time to yield results.}
    \label{tab:results}
\end{table*}

\subsection{Implementation details}
For visual reference prompting methods, the semantic support set $\mathcal{S}^{sem}$ is computed once at the beginning of the evaluation and remains unchanged. The images that constitute $\mathcal{S}^{sem}$ are drawn from the training split of each dataset, while the evaluation is conducted on the validation set when available, or on the test set otherwise.

We adapted visual reference prompting methods by reimplementing them within our codebase, while evaluating open-vocabulary methods using MMSegmentation~\cite{mmseg2020}. Due to the inherent variability in visual prompt selection, we conducted each experiment by randomly sampling five different support sets. We report both the mean and standard deviation of the results. For OVSS methods, we follow their protocol and use a vocabulary containing the dataset classes and, when required, also the \textit{background} class.
For our evaluation metric, we used the standard mean Intersection over Union (mIoU), with the background class excluded from computation when present.

\subsection{Quantitative results}
\Cref{tab:results} reveals that visual prompting methods generally outperform textual approaches across our benchmark, with GFSAM~\cite{zhang2024bridge} achieving the highest average performance (38.7 mIoU on average with five prompts). The performance gap between modalities dramatically varies across domains, with visual methods showing particular strength in specialized contexts while textual approaches remain competitive in common scenes. Increasing visual prompts from one to five consistently improves performance for all methods, with GFSAM~\cite{zhang2024bridge} showing the most substantial average gain (+8.7 mIoU). The high standard deviation observed in some domains (±14.1 for GFSAM~\cite{zhang2024bridge} on Pizza) confirms our hypothesis about prompt selection sensitivity.
Interestingly, the relative performance of prompting modalities dramatically shifts across domains. In common scenes, textual prompts excel on PASCAL VOC~\cite{everingham2010pascal} (CLIP-DINoiser: 62.1 mIoU vs. GFSAM: 45.0 mIoU), yet this advantage reverses on ADE20K~\cite{zhou2017scene} where visual prompting slightly prevails. This pattern contrasts sharply in specialized domains like waste recognition, where GFSAM (52.5 mIoU) outperforms ProxyCLIP (21.1 mIoU) by an impressive 31.4 points on the Trash dataset, and tools, where Matcher achieves an exceptional 88.9 mIoU on Toolkits versus NACLIP's 12.9 mIoU. The urban domain falls between these extremes, with visual methods maintaining a moderate but consistent advantage on both Cityscapes~\cite{cordts2016cityscapes} (GFSAM: 44.6 mIoU vs. ProxyCLIP: 40.1 mIoU) and UAVid~\cite{lyu2020uavid} (SINE: 36.9 mIoU vs. ProxyCLIP: 30.2 mIoU).
Perhaps, the most revealing is the parts domain, where the performance gap varies dramatically between datasets: visual methods vastly outperform textual approaches on House-Parts~\cite{house-dataset} (Matcher: 25.0 mIoU vs. TCL: 7.8 mIoU), yet the difference nearly vanishes on MHPv1~\cite{li2017multiple} (GFSAM: 28.6 mIoU vs. ProxyCLIP: 28.1 mIoU). This suggests textual descriptions adequately capture human anatomy but struggle with architectural elements that benefit from visual reference. Similarly, on PIDray's~\cite{zhang2023pidray} X-ray scans, GFSAM (21.0 mIoU) triples the performance of textual methods (7.5 mIoU), demonstrating textual prompts' inability to capture the unique visual characteristics of tools in security screening contexts.

Our results further validate our strategic emphasis on training-free methods, as trained approaches like SINE~\cite{liu2024simple} and TCL~\cite{cha2023learning} exhibit limited cross-domain adaptability. Despite SINE's~\cite{liu2024simple} strong performance on Cityscapes~\cite{cordts2016cityscapes} (44.1 mIoU), it performs poorly on specialized domains like waste (14.4 mIoU and 8.1 mIoU) and food (19.5 mIoU and 2.7 mIoU). This stark contrast highlights the superior adaptability of training-free methods across diverse visual domains, particularly when leveraging visual prompts that can capture domain-specific visual characteristics that textual descriptions often fail to convey.

\subsection{Qualitative results}
\begin{figure*}[ht]
  \centering
  \includegraphics[width=\linewidth]{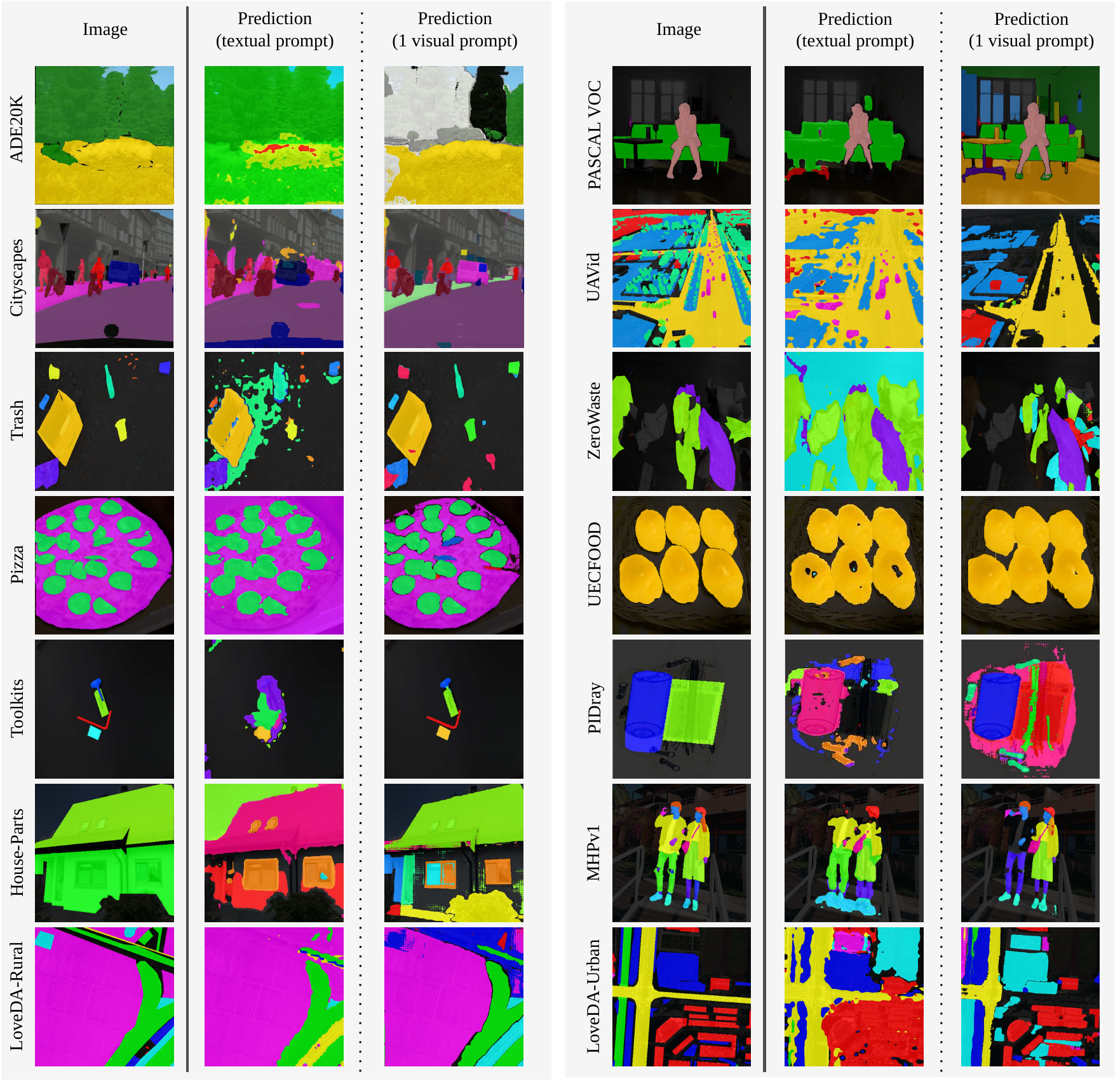}
  \caption{Qualitative results of ProxyCLIP~\cite{lan2024proxyclip} (\textit{textual prompt}) and GFSAM~\cite{zhang2024bridge} (\textit{visual prompt}) across all the dataset in our benchmark.}
  \label{fig:qualitative}
\end{figure*}

In \Cref{fig:qualitative}, we compare the segmentation prediction obtained on our benchmark via textual prompts and a single visual prompts using ProxyCLIP~\cite{lan2024proxyclip} and GFSAM~\cite{zhang2024bridge}.

Segmentation masks generated using visual prompts demonstrate superior boundary refinement and shape definition compared to those produced with textual prompts. This phenomenon is particularly pronounced in the Trash~\cite{trash-dataset} and Toolkits~\cite{toolkit-dataset} datasets, where visual prompts effectively capture intricate object details, such as the cardboard (\textit{yellow class, Trash}) and the Allen key (\textit{red class, Toolkits}), with significantly higher precision than their textual counterparts.
In more challenging datasets such as PIDray~\cite{zhang2023pidray} and MHPv1~\cite{li2017multiple}, both prompting approaches demonstrate limitations in generating comprehensive masks for the specified classes. Although visual prompting successfully segments the sprayer in the X-ray scans of the PIDray~\cite{zhang2023pidray} dataset (\textit{blue class}), it exhibits reduced efficacy in delineating the various body parts in the MHPv1~\cite{li2017multiple} dataset.
Conversely, in datasets such as ADE20K~\cite{zhou2017scene}, Pizza~\cite{pizza_dataset}, UECFOOD~\cite{ege2019new}, Cityscapes~\cite{cordts2016cityscapes}, and PASCAL VOC~\cite{everingham2010pascal}, both methods produce accurate  results. However, in PASCAL VOC~\cite{everingham2010pascal}, the limitations of a single prompt become evident: despite accurate segmentation of target classes, the final prediction contains extraneous classes not present in the image.
A significant qualitative observation emerges from the UAVid~\cite{lyu2020uavid} dataset. In this instance, textual prompts generate an approximate segmentation of the entire scene, whereas visual prompting demonstrates reduced efficacy in capturing numerous buildings and distant vegetation. This observation reveals a fundamental limitation of visual prompting: its diminished capacity to effectively segment small objects in complex, cluttered environments. Textual prompts, despite their inherent imprecision, provide more comprehensive segmentation in such scenarios.

\begin{table}[ht]
    \setlength{\tabcolsep}{1.5pt}
    \renewcommand{\arraystretch}{1.3}   
    \centering
    \resizebox{\columnwidth}{!}{
    \begin{tabular}{lcccccc} 
        \toprule
          & \multicolumn{2}{c}{ADE20K} & \multicolumn{2}{c}{Cityscapes} & \multicolumn{2}{c}{PIDray} \\
         & \#\textit{Forward} & Time & \#\textit{Forward} & Time & \#\textit{Forward} & Time \\ 
        \midrule
        CLIP-DINoiser \cite{wysoczanska2024clip} & 1 & {\footnotesize $\sim$} 0.08s & 1 & {\footnotesize $\sim$} 0.1s & 1 & {\footnotesize $\sim$} 0.07s \\
        NACLIP \cite{hajimiri2024pay}            & 1 & {\footnotesize $\sim$} 0.05s & 1 & {\footnotesize $\sim$} 0.2s & 1 & {\footnotesize $\sim$} 0.05s \\
        ProxyCLIP \cite{lan2024proxyclip}        & 1 & {\footnotesize $\sim$} 0.6s & 1 & {\footnotesize $\sim$} 0.5s & 1 & {\footnotesize $\sim$} 0.2s \\
        
        \midrule
        PerSAM \cite{zhang2023personalize} & 150 & {\footnotesize $\sim$} 320s & 19 & {\footnotesize $\sim$} 40s & 12 & {\footnotesize $\sim$} 25s \\
        Matcher \cite{liu2023matcher} & 150 & {\footnotesize $\sim$} 1,800s & 19 & {\footnotesize $\sim$} 240s & 12 & {\footnotesize $\sim$} 150s \\
        GFSAM \cite{zhang2024bridge} & 150 & {\footnotesize $\sim$} 220s & 19 & {\footnotesize $\sim$} 30s & 12 & {\footnotesize $\sim$} 19s \\ 
        \bottomrule
    \end{tabular}}
    \caption{Inference time and the number of forward required by visual reference (1 prompt) and open-vocabulary methods to produce a segmentation mask for a \textbf{single image}. Inference time is measured on single NVIDIA L4.}
    \label{tab:computation}
\end{table}

\subsection{Computational analysis}
\label{subsec:computation}

\Cref{tab:computation} presents the number of forward passes required by each method to generate a segmentation mask for all classes in a given image, along with the corresponding inference time (in seconds) per image. Open-vocabulary methods (upper part of the table) only require a single forward pass and produce segmentation masks in less than a second. CLIP-DINoiser~\cite{wysoczanska2024clip} and NACLIP~\cite{hajimiri2024pay} demonstrate particularly efficient performance, with inference times below 0.2 seconds per image. Conversely, ProxyCLIP~\cite{lan2024proxyclip}, which achieves superior performance on average, exhibits longer inference times ranging from 0.2 to 0.6 seconds. This increased computational demand can be attributed to its integration of DINO~\cite{caron2021emerging, oquab2023dinov2}.

As detailed in \Cref{subsec:adapting}, visual reference methods (bottom part of the table) requires $n$ forward passes, where $n$ represents the number of classes. Consequently, inference times for these methods are substantially larger than textual prompting and range from 8 to over 1800 seconds per image. Matcher~\cite{liu2023matcher} exhibits the highest computational demand among visual reference methods, with inference times exceeding 30 minutes on ADE20K~\cite{zhou2017scene}, primarily due to its dependence on SAM's computationally intensive automatic mask generator~\cite{kirillov2023segment}. In contrast, GFSAM~\cite{zhang2024bridge} demonstrates the most efficient performance, with an inference time of 220 seconds on ADE20K~\cite{zhou2017scene}.

\section{Conclusions} \label{sec:conclusion}
We presented \method\ (\short), a comprehensive benchmark evaluating both visual and textual prompts for semantic segmentation across 14 datasets spanning 7 diverse domains. Our experiments with 5 open-vocabulary and 4 visual reference prompt methods revealed distinct strengths and limitations of each prompting modality.

Open-vocabulary methods excelled in common scenes and urban environments where concepts are readily described textually. ProxyCLIP~\cite{lan2024proxyclip} variants achieved the highest scores on PASCAL VOC~\cite{everingham2010pascal}, ADE20K~\cite{zhou2017scene}, and Cityscapes~\cite{cordts2016cityscapes}. However, these methods performed poorly in specialized domains like tools and parts, where textual descriptions inadequately capture complex visual characteristics.
Visual reference prompt methods demonstrated more consistent performance across domains, with GFSAM~\cite{zhang2024bridge} showing strong results when provided with multiple visual prompts. These methods performed very well in specialized categories like tools (88.9\% on Toolkits) where visual examples offer more effective guidance than text descriptions. Their performance, however, varied considerably depending on the quality and representativeness of prompts, revealing a significant sensitivity to prompt selection.

\myparagraph{Open challenges}
Our benchmark identifies several key challenges for future research.
\begin{itemize}
  \item \textbf{Computational efficiency:} Visual reference prompt methods process binary class-by-class, resulting in slower inference compared to open-vocabulary approaches.
  \item \textbf{Prompt selection sensitivity:} Visual in-context learning methods show high sensitivity to prompt selection, undermining their reliability in practical applications.
  \item \textbf{Domain specialization:} Open-vocabulary methods struggle with specialized domains due to poor text-to-visual feature alignment.
\end{itemize}

\myparagraph{Acknowledgements}
This publication is part of the project PNRR-NGEU which has received funding from the MUR – DM 117/2023. We acknowledge ISCRA for awarding this project access to the LEONARDO supercomputer, owned by the EuroHPC Joint Undertaking, hosted by CINECA (Italy).
{
  \small
  \bibliographystyle{ieeenat_fullname}
  \bibliography{main}
}

\clearpage
\maketitlesupplementary

\section{Semantic support set generation}

In \Cref{alg:semanticsupportset} we describe in detail how the semantic support set $\mathcal{S}^{sem}$ has been generated.

\renewcommand{\algorithmicrequire}{\textbf{Input:}}
\renewcommand{\algorithmicensure}{\textbf{Output:}}
\begin{algorithm}
\caption{Generation of Semantic Support Set}
\label{alg:semanticsupportset}
\begin{algorithmic}[1]
\Require Set of class IDs $\mathcal{C}$, Set of training images names for each class $\mathcal{D}_c$, Number of visual prompts $k$
\Ensure Semantic support set $\mathcal{S}^{sem}$
\State $\mathcal{S}^{sem} \gets \emptyset$
\For{each $c \in \mathcal{C}$} \Comment{Iterate over class IDs}
    \State $\mathcal{I}_s \gets \emptyset$ \Comment{Support images}
    \State $\mathcal{M}_s \gets \emptyset$ \Comment{Support masks}
    \State $\mathcal{N}_s \gets \emptyset$ \Comment{Support names}
    \While{$|\mathcal{I}_s| < k$}
        \State Sample $n_s \sim \textit{Uniform}(\mathcal{D}_c \setminus \mathcal{N}_s)$ 
        \State $i_s \gets \texttt{LoadImage}(n_s)$
        \State $m_s \gets \texttt{LoadMask}(n_s)$
        \State $m_s \gets \mathds{1}(m_s = c)$ \Comment{Select only the class $c$}

        \State $\mathcal{I}_s \gets \mathcal{I}_s \cup \{i_s\}$
        \State $\mathcal{M}_s \gets \mathcal{M}_s \cup \{m_s\}$
        \State $\mathcal{N}_s \gets \mathcal{N}_s \cup \{n_s\}$ 
    \EndWhile
    \State $\mathcal{S}^{sem} \gets \mathcal{S} \cup \{(\mathcal{I}_s, \mathcal{M}_s, \mathcal{N}_s)\}$
\EndFor
\end{algorithmic}
\end{algorithm}

\section{Additional implementation details}
\myparagraph{Visual reference prompt methods.} Following the original implementations, visual reference prompt methods are evaluated using DINOv2~\cite{oquab2023dinov2} with ViT-L/14~\cite{dosovitskiy2020image} (when applicable), and SAM~\cite{kirillov2023segment} with ViT-H~\cite{dosovitskiy2020image} (when applicable).

\myparagraph{Open-vocabulary methods.} To ensure a fair comparison, we report the results for open-vocabulary methods without applying any mask refinement step (\eg PAMR~\cite{araslanov2020single}). Furthermore, we use ViT-L/14~\cite{dosovitskiy2020image} as the visual backbone for NACLIP~\cite{hajimiri2024pay} and ProxyCLIP~\cite{lan2024proxyclip}, while all other methods are evaluated using ViT-B/16~\cite{dosovitskiy2020image}.

\section{Dataset classes}

For each dataset comprised in our \short\ benchmark, we report the list of classes.

\myparagraph{ADE20K} The ADE20K~\cite{zhou2017scene} dataset is made up of 150 classes. The classes are: \texttt{wall}, \texttt{building}, \texttt{sky}, \texttt{floor}, \texttt{tree}, \texttt{ceiling}, \texttt{road}, \texttt{bed}, \texttt{windowpane}, \texttt{grass}, \texttt{cabinet}, \texttt{sidewalk}, \texttt{person}, \texttt{earth}, \texttt{door}, \texttt{table}, \texttt{mountain}, \texttt{plant}, \texttt{curtain}, \texttt{chair}, \texttt{car}, \texttt{water}, \texttt{painting}, \texttt{sofa}, \texttt{shelf}, \texttt{house}, \texttt{sea}, \texttt{mirror}, \texttt{rug}, \texttt{field}, \texttt{armchair}, \texttt{seat}, \texttt{fence}, \texttt{desk}, \texttt{rock}, \texttt{wardrobe}, \texttt{lamp}, \texttt{bathtub}, \texttt{railing}, \texttt{cushion}, \texttt{base}, \texttt{box}, \texttt{column}, \texttt{signboard}, \texttt{chestofdrawers}, \texttt{counter}, \texttt{sand}, \texttt{sink}, \texttt{skyscraper}, \texttt{fireplace}, \texttt{refrigerator}, \texttt{grandstand}, \texttt{path}, \texttt{stairs}, \texttt{runway}, \texttt{case}, \texttt{pooltable}, \texttt{pillow}, \texttt{screendoor}, \texttt{stairway}, \texttt{river}, \texttt{bridge}, \texttt{bookcase}, \texttt{blind}, \texttt{coffeetable}, \texttt{toilet}, \texttt{flower}, \texttt{book}, \texttt{hill}, \texttt{bench}, \texttt{countertop}, \texttt{stove}, \texttt{palm}, \texttt{kitchenisland}, \texttt{computer}, \texttt{swivelchair}, \texttt{boat}, \texttt{bar}, \texttt{arcademachine}, \texttt{hovel}, \texttt{bus}, \texttt{towel}, \texttt{light}, \texttt{truck}, \texttt{tower}, \texttt{chandelier}, \texttt{awning}, \texttt{streetlight}, \texttt{booth}, \texttt{televisionreceiver}, \texttt{airplane}, \texttt{dirttrack}, \texttt{apparel}, \texttt{pole}, \texttt{land}, \texttt{bannister}, \texttt{escalator}, \texttt{ottoman}, \texttt{bottle}, \texttt{buffet}, \texttt{poster}, \texttt{stage}, \texttt{van}, \texttt{ship}, \texttt{fountain}, \texttt{conveyerbelt}, \texttt{canopy}, \texttt{washer}, \texttt{plaything}, \texttt{swimmingpool}, \texttt{stool}, \texttt{barrel}, \texttt{basket}, \texttt{waterfall}, \texttt{tent}, \texttt{bag}, \texttt{minibike}, \texttt{cradle}, \texttt{oven}, \texttt{ball}, \texttt{food}, \texttt{step}, \texttt{tank}, \texttt{tradename}, \texttt{microwave}, \texttt{pot}, \texttt{animal}, \texttt{bicycle}, \texttt{lake}, \texttt{dishwasher}, \texttt{screen}, \texttt{blanket}, \texttt{sculpture}, \texttt{hood}, \texttt{sconce}, \texttt{vase}, \texttt{trafficlight}, \texttt{tray}, \texttt{ashcan}, \texttt{fan}, \texttt{pier}, \texttt{crtscreen}, \texttt{plate}, \texttt{monitor}, \texttt{bulletinboard}, \texttt{shower}, \texttt{radiator}, \texttt{glass}, \texttt{clock}, and \texttt{flag}.

\myparagraph{PASCAL VOC 2012} The PASCAL VOC 2012~\cite{everingham2010pascal} dataset is made up of 21 classes. The classes are: \texttt{background}, \texttt{aeroplane}, \texttt{bicycle}, \texttt{bird}, \texttt{boat}, \texttt{bottle}, \texttt{bus}, \texttt{car}, \texttt{cat}, \texttt{chair}, \texttt{cow}, \texttt{diningtable}, \texttt{dog}, \texttt{horse}, \texttt{motorbike}, \texttt{person}, \texttt{pottedplant}, \texttt{sheep}, \texttt{sofa}, \texttt{train}, and \texttt{tvmonitor}.

\myparagraph{Cityscapes} The Cityscapes~\cite{cordts2016cityscapes} dataset is composed of 19 classes. The classes are: \texttt{road}, \texttt{sidewalk}, \texttt{building}, \texttt{wall}, \texttt{fence}, \texttt{pole}, \texttt{trafficlight}, \texttt{trafficsign}, \texttt{vegetation}, \texttt{terrain}, \texttt{sky}, \texttt{person}, \texttt{rider}, \texttt{car}, \texttt{truck}, \texttt{bus}, \texttt{train}, \texttt{motorcycle} and \texttt{bicycle}.

\myparagraph{UAVid} The UAVid~\cite{lyu2020uavid} dataset is made up of 7 classes. The classes are: \texttt{Building}, \texttt{Road}, \texttt{Static Car}, \texttt{Tree}, \texttt{Vegetation}, \texttt{Human} and \texttt{Moving Car}. A background class is also present in the dataset.

\myparagraph{Trash} The Trash~\cite{trash-dataset} dataset is made up of 12 classes. The classes are: \texttt{Aluminium foil}, \texttt{Cigarette}, \texttt{Clear plastic bottle}, \texttt{Corrugated carton}, \texttt{Disposable plastic cup}, \texttt{Drink Can}, \texttt{Egg Carton}, \texttt{Foam cup}, \texttt{Food Can}, \texttt{Garbage bag}, \texttt{Glass bottle}, \texttt{Glass cup}, \texttt{Metal bottle cap}, \texttt{Other carton}, \texttt{Other plastic bottle}, \texttt{Paper cup}, \texttt{Plastic bag - wrapper}, \texttt{Plastic bottle cap}, \texttt{Plastic lid}, \texttt{Plastic straw}, \texttt{Pop tab} and \texttt{Styrofoam piece}. A background class is also present in the dataset.

\myparagraph{ZeroWaste} The ZeroWaste~\cite{bashkirova2022zerowaste} dataset is made up of 4 classes. The classes are: \texttt{rigid plastic}, \texttt{cardboard}, \texttt{metal} and \texttt{soft plastic}. A background class is also present in the dataset.    

\myparagraph{Pizza} The Pizza~\cite{pizza_dataset} dataset is made up of 5 classes. The classes are: \texttt{Mushroom}, \texttt{Pepper}, \texttt{Pepperoni}, \texttt{Tomato} and \texttt{pizza}. A background class is also present in the dataset.

\myparagraph{UECFOOD} The UECFOOD~\cite{ege2019new} dataset is made up of 102 classes. The classes are: \texttt{rice}, \texttt{eels on rice}, \texttt{pilaf}, \texttt{chicken-'n'-egg on rice}, \texttt{pork cutlet on rice}, \texttt{beef curry}, \texttt{sushi}, \texttt{chicken rice}, \texttt{fried rice}, \texttt{tempura bowl}, \texttt{bibimbap}, \texttt{toast}, \texttt{croissant}, \texttt{roll bread}, \texttt{raisin bread}, \texttt{chip butty}, \texttt{hamburger}, \texttt{pizza}, \texttt{sandwiches}, \texttt{udon noodle}, \texttt{tempura udon}, \texttt{soba noodle}, \texttt{ramen noodle}, \texttt{beef noodle}, \texttt{tensin noodle}, \texttt{fried noodle}, \texttt{spaghetti}, \texttt{Japanese-style pancake}, \texttt{takoyaki}, \texttt{gratin}, \texttt{sauteed vegetables}, \texttt{croquette}, \texttt{grilled eggplant}, \texttt{sauteed spinach}, \texttt{vegetable tempura}, \texttt{miso soup}, \texttt{potage}, \texttt{sausage}, \texttt{oden}, \texttt{omelet}, \texttt{ganmodoki}, \texttt{jiaozi}, \texttt{stew}, \texttt{teriyaki grilled fish}, \texttt{fried fish}, \texttt{grilled salmon}, \texttt{salmon meuniere}, \texttt{sashimi}, \texttt{grilled pacific saury}, \texttt{sukiyaki}, \texttt{sweet and sour pork}, \texttt{lightly roasted fish}, \texttt{steamed egg hotchpotch}, \texttt{tempura}, \texttt{fried chicken}, \texttt{sirloin cutlet}, \texttt{nanbanzuke}, \texttt{boiled fish}, \texttt{seasoned beef with potatoes}, \texttt{hambarg steak}, \texttt{beef steak}, \texttt{dried fish}, \texttt{ginger pork saute}, \texttt{spicy chili-flavored tofu}, \texttt{yakitori}, \texttt{cabbage roll}, \texttt{rolled omelet}, \texttt{egg sunny-side up}, \texttt{fermented soybeans}, \texttt{cold tofu}, \texttt{egg roll}, \texttt{chilled noodle}, \texttt{stir-fried beef and peppers}, \texttt{simmered pork}, \texttt{boiled chicken and vegetables}, \texttt{sashimi bowl}, \texttt{sushi bowl}, \texttt{fish-shaped pancake with bean jam}, \texttt{shrimp with chill source}, \texttt{roast chicken}, \texttt{steamed meat dumpling}, \texttt{omelet with fried rice}, \texttt{cutlet curry}, \texttt{spaghetti meat sauce}, \texttt{fried shrimp}, \texttt{potato salad}, \texttt{green salad}, \texttt{macaroni salad}, \texttt{Japanese tofu and vegetable chowder}, \texttt{pork miso soup}, \texttt{chinese soup}, \texttt{beef bowl}, \texttt{kinpira-style sauteed burdock}, \texttt{rice ball}, \texttt{pizza toast}, \texttt{dipping noodles}, \texttt{hot dog}, \texttt{french fries}, \texttt{mixed rice}, \texttt{goya chanpuru}, \texttt{others} and \texttt{beverage}. A background class is also present in the dataset.

\myparagraph{Toolkits} The Toolkits~\cite{toolkit-dataset} dataset is made up of 8 classes. The classes are: \texttt{Allen-key}, \texttt{block}, \texttt{gasket}, \texttt{plier}, \texttt{prism}, \texttt{screw}, \texttt{screwdriver} and \texttt{wrench}. A background class is also present in the dataset.

\myparagraph{PIDray} The PIDray~\cite{zhang2023pidray} dataset is made up of 12 classes. The classes are: \texttt{Baton}, \texttt{Pliers}, \texttt{Hammer}, \texttt{Powerbank}, \texttt{Scissors}, \texttt{Wrench}, \texttt{Gun}, \texttt{Bullet}, \texttt{Sprayer}, \texttt{HandCuffs}, \texttt{Knife} and \texttt{Lighter}. A background class is also present in the dataset.

\myparagraph{House-Parts} The House-Parts~\cite{house-dataset} dataset is made up of 22 classes. The classes are: \texttt{aluminium door}, \texttt{aluminium window}, \texttt{cellar window}, \texttt{mint cond roof}, \texttt{plaster}, \texttt{plastic door}, \texttt{plastic window}, \texttt{plate fascade}, \texttt{wooden door}, \texttt{wooden fascade}, \texttt{wooden window} and \texttt{worn cond roof}. A background class is also present in the dataset.

\myparagraph{MHPv1} The MHPv1~\cite{li2017multiple} dataset is made up of 17 classes. The classes are: \texttt{hat}, \texttt{hair}, \texttt{sunglasses}, \texttt{upper clothes}, \texttt{skirt}, \texttt{pants}, \texttt{dress}, \texttt{belt}, \texttt{left shoe}, \texttt{right shoe}, \texttt{face}, \texttt{left leg}, \texttt{right leg}, \texttt{left arm}, \texttt{right arm}, \texttt{bag} and \texttt{scarf}. A background class is also present in the dataset.

\myparagraph{LoveDA} The LoveDA~\cite{wang2021loveda} dataset, which is composed by LoveDA-Rural and LoveDA-Urban, is made up of 6 classes. The classes are: \texttt{building}, \texttt{road}, \texttt{water}, \texttt{barren}, \texttt{forest} and \texttt{agriculture}. A background class is also present in the dataset.

\end{document}